\documentclass[letterpaper]{article} 
\usepackage{aaai2026}
\usepackage{times}  
\usepackage{helvet}  
\usepackage{courier}  
\usepackage[hyphens]{url}  
\usepackage{graphicx} 
\usepackage{natbib}  
\usepackage{caption} 
\usepackage{algorithm}
\usepackage{algorithmic}

\usepackage{bm}
\usepackage{amssymb}
\usepackage{amsmath}
\usepackage{booktabs}
\usepackage{xcolor} 
\usepackage{cancel}
\usepackage{multirow}
\usepackage{array}
\usepackage{makecell}
\usepackage{pifont}

\usepackage{newfloat}
\usepackage{listings}
\DeclareCaptionStyle{ruled}{labelfont=normalfont,labelsep=colon,strut=off} 
\floatstyle{ruled}
\newfloat{listing}{tb}{lst}{}
\floatname{listing}{Listing}
\nocopyright

\title{Causality Matters: How Temporal Information \\ Emerges in Video Language Models}
\author {
    Yumeng Shi, 
    Quanyu Long, 
    Yin Wu, 
    Wenya Wang
}
\affiliations {
    Nanyang Technological University \\
    yumeng001@e.ntu.edu.sg, quanyu001@e.ntu.edu.sg, wuyi0023@e.ntu.edu.sg, wangwy@ntu.edu.sg
}

\usepackage{bibentry}

\begin{document}

\maketitle

\begin{abstract}
Video language models (VideoLMs) have made significant progress in multimodal understanding. However, temporal understanding, which involves identifying event order, duration, and relationships across time, still remains a core challenge. Prior works emphasize positional encodings (PEs) as a key mechanism for encoding temporal structure. Surprisingly, we find that removing or modifying PEs in video inputs yields minimal degradation in the performance of temporal understanding. In contrast, reversing the frame sequence while preserving the original PEs causes a substantial drop. To explain this behavior, we conduct substantial analysis experiments to trace how temporal information is integrated within the model. We uncover a causal information pathway: temporal cues are progressively synthesized through inter-frame attention, aggregated in the final frame, and subsequently integrated into the query tokens. This emergent mechanism shows that temporal reasoning emerges from inter-visual token interactions under the constraints of causal attention, which implicitly encodes temporal structure. Based on these insights, we propose two efficiency-oriented strategies: staged cross-modal attention and a temporal exit mechanism for early token truncation. Experiments on two benchmarks validate the effectiveness of both approaches. To the best of our knowledge, this is the first systematic study of video temporal understanding in VideoLMs, offering insights for future model improvement. Our code is available at \textit{https://github.com/ANDgate99/Causality-Matters}.
\end{abstract}


\section{Introduction}
Video language models (VideoLMs)~\cite{zhang2024longva, lin2023video}, built on the foundation of large language models (LLMs)~\cite{openai2024gpt4o, yang2025qwen3}, have achieved significant progress in a wide range of video understanding tasks, including captioning, question answering, and temporal reasoning. Among the various challenges posed by these tasks, temporal understanding~\cite{cai2024temporalbench, shangguantomato}, defined as the ability to recognize and interpret the order, duration, and relationships between events over time, stands out as both fundamental and difficult. From reasoning about causal relationships in instructional videos to maintaining narrative coherence in long-form content, temporal understanding plays a vital role in enabling intelligent video-language interactions.

\begin{figure}[t]
    \centering
    \includegraphics[width=1\columnwidth]{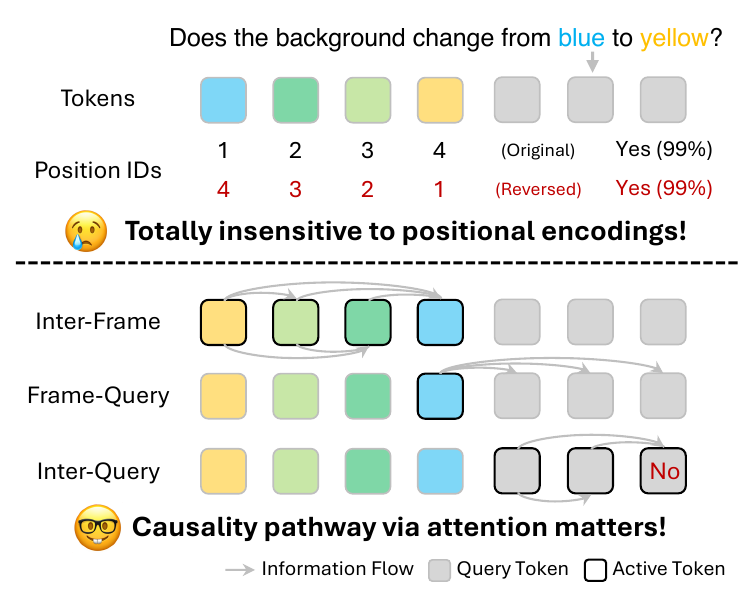}
    \caption{Temporal information emerges through a causal attention pathway instead of PEs. Reversing position IDs has little effect, but reversing the frame order while keeping position IDs aligned with the original order changes the output.}
    \label{fig:intro}
\end{figure}

In the pursuit of enhancing the temporal understanding capabilities of VideoLMs~\cite{litemporal, nguyen2025moose}, many studies~\cite{liu2025vrope} have focused on what is likely the model's most direct method for modeling temporal structures: the positional encodings (PEs). It is commonly assumed that using a more sophisticated positional encoding function leads directly to improved temporal awareness. This belief has spurred a variety of modifications, from extending PEs to higher dimensions~\cite{Qwen2.5-VL, weivideorope} to designing variable-rate formats~\cite{ge2024v2pe} to better capture temporal information. However, this intense focus on an explicit signal raises a basic but often overlooked question: \textbf{\ding{172} \textit{To what extent do PEs support temporal understanding in modern VideoLMs?}}

To better understand the role of PEs, our analysis begins by iteratively removing them layer by layer and selectively modifying them in the video and query inputs separately. The outcome is revealing: PEs have only a marginal impact on temporal understanding performance beyond the first layer, and while query PEs play a more significant role, video input PEs contribute little. These observations raised a more fundamental question: \textbf{\ding{173} \textit{If the model does not primarily rely on explicit PEs for temporal understanding, what is the primary source of this capability?}} 

This question motivates a follow-up experiment, where we evaluate the models on deliberately reversed video input sequences while preserving PEs. In contrast to the PE ablation, this reversal severely degrades the performance. These initial findings suggest that VideoLMs are highly sensitive to the order of video frames, yet do not primarily rely on explicit PEs to detect it. Building on this, we propose a new hypothesis: Currently, temporal understanding is not a fixed attribute extracted from PEs, but an emergent phenomenon derived from the order-aware processing imposed by causal masking in attention. More specifically, it emerges from: \textbf{\ding{174} \textit{How the causal attention mechanism permits temporal information to be generated and to flow from interaction between tokens across layers.}}

We examine the hypothesis by tracing the model's temporal understanding process backward from its final output. This reveals that the model generates its answer primarily from the query, rather than directly referencing the video evidence at the final stage. Tracing this dependency further upstream, we observe a multi-stage process in the model's early and middle layers: raw information is first aggregated across the video frames, and the resulting representation is then used to enrich the query's context. In addition, we validate that this entire information pathway is not merely correlational but genuinely causal, providing direct evidence of how temporal understanding emerges within the model.

Based on experiments with the advanced Qwen2.5-VL~\cite{Qwen2.5-VL} and LLaVA-OneVision~\cite{li2024llava}, which represent state-of-the-art VideoLMs, we identify several key takeaways about how modern VideoLMs internally process temporal information, as shown in Figure~\ref{fig:intro}:
\begin{itemize}
    \item[\ding{172}] PEs are not the primary source of temporal information currently, suggesting emphasis may shift from PE design to how models leverage them during training.
    \item[\ding{173}] Temporal information emerges from the causal attention mechanism's order-aware structure, highlighting the potential of sequential architectures for temporal modeling.
    \item[\ding{174}] Temporal information is progressively constructed via a multi-stage causal pathway: it emerges from long-range inter-video interactions, flows into the query through the final video frame where it is aggregated, and is then processed independently by the query without further contribution from the video input.
\end{itemize}

Building on these insights, we discuss two potential application scenarios: (1) \textit{Staged Modality Interaction for Sparse Attention}, which reduces cross-modal interaction to improve computational efficiency; (2) \textit{Temporal Exit for KV Cache Compression}, which alleviates GPU memory pressure by discarding tokens that no longer contribute to temporal information propagation. We demonstrate the feasibility of both approaches through experiments on two datasets.

\section{Related Work}
\subsection{Video Language Model}
With the rise of LLMs~\cite{openai2024gpt4o, bi2024deepseek}, numerous studies~\cite{kangaroogroup, zhang2024longva, yao2024minicpm, damonlpsg2025videollama3} have explored their integration into VideoLMs, which increasingly outperform traditional vision-only methods. One major direction focuses on enhancing the video encoder~\cite{zhao2024videoprism, choudhury2024don, chungunifying}. Another line of work~\cite{chen2024internvl, weivideorope, ge2024v2pe, Qwen2.5-VL} adapts LLMs to better accommodate multimodal inputs. In addition, other studies aim to improve the efficiency of processing long video inputs~\cite{fu2024framefusion, he2024ma, shi2025static}. Collectively, these advancements have driven progress in video-language understanding.

\subsection{Temporal Understanding}

VideoLMs face unique challenges in video temporal understanding, which requires detecting similarities and differences across frames. Benchmarks~\cite{cai2024temporalbench, shangguantomato, plizzari2025omnia, liu2024tempcompass} have been proposed to evaluate this capability. Besides, various approaches~\cite{nie2024slowfocus, hu2024enhancing, zhao2025videoexpert, fateh2024video} have been introduced to improve this capability. Among them, T3~\cite{litemporal} transfers temporal skills from synthetic text tasks to VideoLMs. Despite these advances, the internal mechanisms by which VideoLMs capture temporal information remain underexplored. Understanding how temporal information is extracted is essential for building more robust and generalizable models.

\subsection{Model Mechanistic Interpretability}

Many studies analyze how LLMs make predictions~\cite{zhao2024towards, goldshmidt2024tokenshap, biran2024hopping, wang2023label}. This line of research has also been extended to multimodal large language models~\cite{yu2024understanding, golovanevsky2025vlms, zhang2025cross}. For instance, \citet{basu2024understanding} uses causal tracing to reveal how early-layer modules capture and transmit visual information. However, most existing work focuses on single-image tasks, and the interpretability of VideoLMs, particularly for temporal reasoning, remains underexplored. This gap motivates our work, which seeks to explain how temporal information is captured within VideoLMs.



\section{Experimental Setting}
\begin{table}[t]
\small
\centering
\begin{tabular}{llc}
    \toprule
    \textbf{Format} & \textbf{Description} & \textbf{Count} \\
    \midrule
    Yes or No            &  Validate statement correctness.       & 2453 \\
    Multiple Choice      &  Select from multiple options.         & 1580 \\
    Caption Matching     &  Select the aligned caption.           & 1503 \\
    Captioning           &  Generate a caption.                   & 2004 \\
    \bottomrule
\end{tabular}
    \caption{Introduction for task types in TempCompass.}
    \label{tab:dataset}
\end{table}

\subsection{Dataset}
We conduct analysis on TempCompass~\cite{liu2024tempcompass}, which encompasses a diverse range of temporal phenomena, including action, speed, direction, attribute change, and event order. It provides a fine-grained assessment of a model’s ability to reason over temporal information. The benchmark includes four task formats, summarized in Table~\ref{tab:dataset}. Since our evaluation focuses on the model's first predicted token, we convert the caption generation task into a multiple-choice format, where the model is required to select the most temporally accurate option.

To validate our proposed efficiency-oriented strategies, we run downstream application experiments on the multiple-choice subset of the NExT-QA dataset~\cite{xiao2021next}, which emphasizes explanatory reasoning over causal and temporal relationships. The test set contains 8,564 questions. To further assess generalizability, we additionally report results on the open-ended ActivityNet-QA dataset~\cite{caba2015activitynet}, included in the appendix.

\subsection{Model}
We evaluate two representative VideoLMs: Qwen2.5-VL and LLaVA-OneVision. Qwen2.5-VL~\cite{Qwen2.5-VL} uses multi-dimensional positional encodings tailored for video, making it well-suited to study PE impact on spatiotemporal modeling. In contrast, LLaVA-OneVision~\cite{li2024llava} is another state-of-the-art multimodal model that does not adapt PEs for video, allowing us to derive insights that are agnostic to special PE designs. Unless specialized otherwise, we report results using Qwen2.5-VL-7B in the main text. Results for Qwen2.5-VL-3B and LLaVA-OneVision are provided in the appendix.

To facilitate a deeper understanding of VideoLM behavior in our experiments, we briefly outline the common architectural framework of modern VideoLMs.

\subsubsection{Input Construction.}
VideoLMs process inputs by integrating an instruction, visual features, and a query into a unified token sequence. The instruction, providing task-specific guidance, is embedded as $\bm{T}_i \in \mathbb{R}^{L_i \times D}$. The visual input comprises sampled video frames, encoded and projected into the language embedding space as $\bm{V} \in \mathbb{R}^{T \times H \times W \times D}$, where $T$ is the temporal resolution and $H$, $W$ are the spatial dimensions. The tensor is then flattened into $\bm{T}_v \in \mathbb{R}^{THW \times D}$. The user query is embedded as $\bm{T}_q \in \mathbb{R}^{L_q \times D}$. The final model input is the concatenated sequence $\bm{T} = [\bm{T}_i; \bm{T}_v; \bm{T}_q]$, enabling joint multimodal reasoning.

\subsubsection{Feature Interaction.}
The concatenated token sequence $\bm{T}$ is fed into the VideoLM. At each layer, the model computes hidden states $\bm{H} = [\bm{H}_i; \bm{H}_v; \bm{H}_q]$, corresponding to instruction, visual, and query tokens, respectively. Multimodal token interactions are governed by the attention mechanism:
\begin{equation}
\label{eq:1}
\bm{Q} = \bm{W}_Q \bm{H}, \quad
\bm{K} = \bm{W}_K \bm{H}, \quad
\bm{V} = \bm{W}_V \bm{H},
\end{equation}
\begin{equation}
\label{eq:2}
\bm{A} = \text{Softmax} \left(
\frac{\text{PE}(\bm{Q}) \cdot \text{PE}(\bm{K})^\mathrm{T}} {\sqrt{d_k}}
+ \bm{M}_C \right) \bm{V},
\end{equation}
where $\bm{W}_Q$, $\bm{W}_K$, and $\bm{W}_V$ are learnable projection matrices, $d_k$ is the dimensionality of each attention head, and $\bm{M}_C$ is a causal mask used for autoregressive decoding. The operator $\text{PE}(\cdot)$ denotes the position encoding function. The attention output is passed through a feed-forward network with residual connections and layer normalization to produce updated hidden states. At the final layer, a vocabulary distribution is computed to predict the next token.

\subsection{Evaluation}
For analysis experiments, we assess the model’s temporal sensitivity by tracking changes in the predicted probability of the ground-truth token after incorporating certain perturbations:
\begin{equation}
\label{eq:3}
P_C = \bm{\tilde{P}}_{\text{next}} \left(\bm{t}_{\text{gt}}\right) -
\bm{P}_{\text{next}} \left(\bm{t}_{\text{gt}}\right),
\end{equation}
where $\bm{\tilde{P}}_{\text{next}}$ is the output probability distribution under a perturbed setting, and $\bm{P}_{\text{next}}$ is the distribution from the base setting. A larger $P_C$ reflects a stronger influence of the perturbation on temporal reasoning. For downstream application experiments, we report the standard accuracy.

\subsection{Implementation Detail}
Our experiments are implemented using PyTorch with the Transformers library. For video input processing, we uniformly sample 8 frames from each video. For Qwen2.5-VL, its internal frame merging mechanism results in 4 effective frames per video. To ensure a controlled evaluation setting, we constrain the model to generate only a single output token, and we analyze its probability distribution to measure temporal effects. Additional experimental configurations are detailed in relevant sections and the appendix.

\section{Is Positional Encoding the Key to Temporal Understanding?}
\label{sec:PE}
\begin{figure}
    \centering
    \includegraphics[width=1\columnwidth]{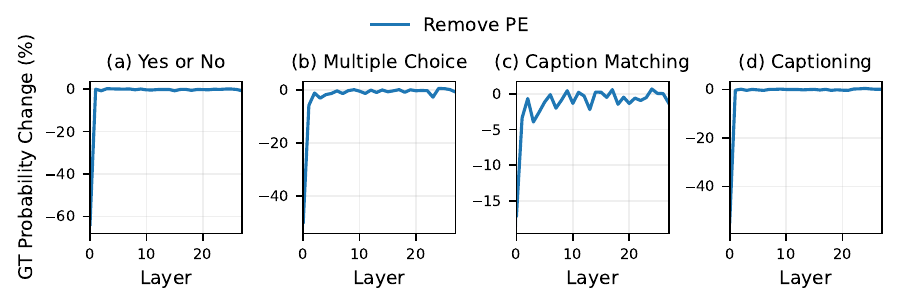}
    \caption{Effect of layer-wise PE ablation. Each point shows the change in ground-truth answer probability when the PE is removed from a layer. An obvious drop appears only at the first layer, while later layers show minimal impact.}
    \label{fig:position}
\end{figure}

Positional encodings (PEs) are widely considered essential for modeling temporal relationships in VideoLMs, especially in recent architectures like Qwen2.5-VL that adopt advanced 3D encodings. The belief that PEs are a primary driver of temporal understanding has spurred research into more sophisticated PE designs. However, it remains unclear whether these encodings are truly the main source of temporal information. In this section, we conduct experiments that isolate the role of PEs and test the hypothesis that temporal information may instead emerge from the causal attention mechanism’s inherent sensitivity to token order.

\subsection{Layer-wise Temporal Sensitivity to PEs}

\subsubsection{Experiment.}
To probe the importance of PEs in temporal information extraction, we first quantify their contribution at different layers of the model. Specifically, we remove the PE terms from Equation~\ref{eq:2} at one layer at a time, leaving all other components unchanged. This allows us to isolate the effect of PEs on temporal modeling at each layer, and to quantify where and how positional signals are most influential.

\subsubsection{Results.}
As shown in Figure~\ref{fig:position}, the model exhibits minimal sensitivity to the removal of PEs across most layers. Across all task types, ablating PEs from intermediate and deeper layers results in minimal performance change, typically within ±2\%. This suggests that these layers are relatively insensitive to the presence of explicit positional signals. A notable exception occurs in the first layer, where removing PEs causes a substantial drop in predicted probability of the correct token, up to 60\%. \textbf{This stark contrast reveals that the impact of PEs for temporal understanding is restricted to the first layer of the model, as later layers exhibit minimal sensitivity to their presence or absence.}


\begin{figure}
    \centering
    \includegraphics[width=0.85\columnwidth]{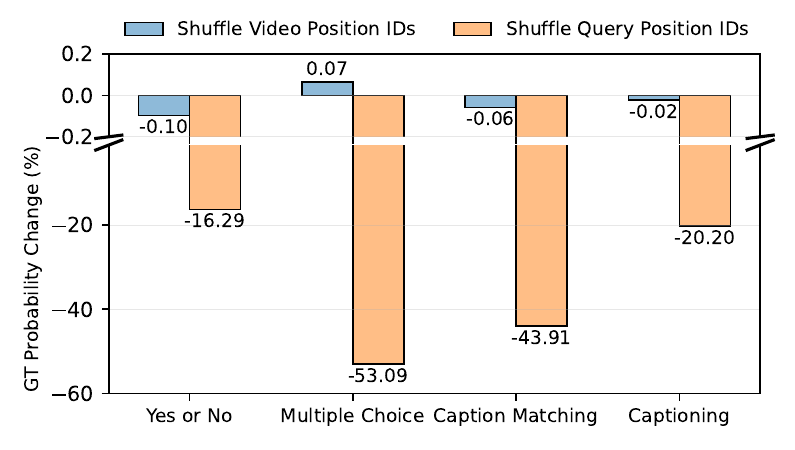}
    \caption{Effect of modality-specific position ID shuffling in the first layer. Shuffling query position IDs leads to significant performance drops, while shuffling video position IDs has minimal effect, highlighting the dominant role of PEs in the textual rather than the intended video modality.}
    \label{fig:shuffle}
\end{figure}

\subsection{Modality-Specific Temporal Roles of PEs}

\subsubsection{Experiment.}
Building on the observed drop in ground-truth probability when PEs are removed from the first layer, we further examine their role in temporal understanding through a targeted, modality-specific experiment. Specifically, we investigate whether the contribution of PEs for temporal understanding differs across input modalities. To this end, we independently shuffle the position IDs of either the video tokens or the query tokens in the first layer, while keeping the other modality unchanged. This setup enables us to directly assess the different impact of PEs on video and textual modalities during temporal modeling.

\subsubsection{Result.}
As shown in Figure~\ref{fig:shuffle}, the results reveal a clear and consistent disparity across modalities. Shuffling the position IDs of query tokens in the first layer significantly degrades performance across all task types. The most pronounced drops occur in the ``Multiple Choice" task, with ground-truth probability decreasing by 53.09\%. In contrast, shuffling the position IDs of video tokens leads to negligible effects. For instance, the performance decreases by only 0.02\% in ``Captioning" and even increases slightly by 0.07\% in ``Multiple Choice", suggesting that the model is largely insensitive to positional distortions in the video modality at this stage. \textbf{Although special PEs are designed to capture temporal structure in video inputs, their impact appears to be concentrated in the textual modality.}


\subsection{PE vs. Order in Temporal Modeling}
\begin{figure}
    \centering
    \includegraphics[width=0.85\columnwidth]{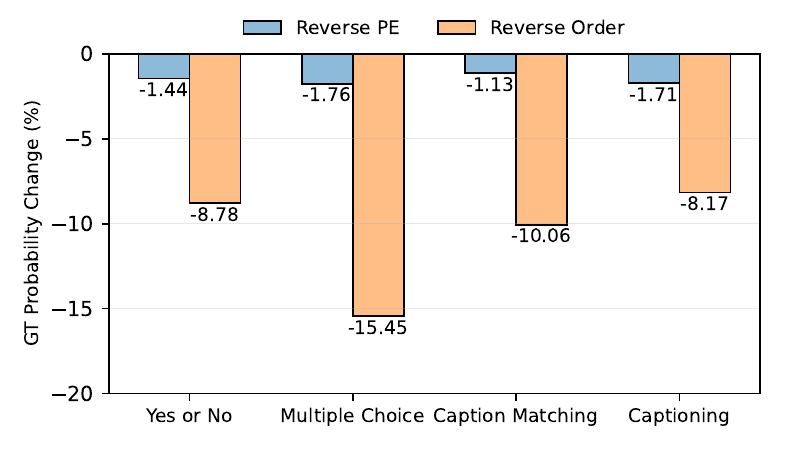}
    \caption{Effect of reversing position IDs versus frame order. Reversing the frame order results in significantly larger performance drops across all task types, suggesting that temporal understanding is primarily driven by the inherent order of frames rather than by positional encodings.}
    \label{fig:ablate}
\end{figure}

\subsubsection{Experiment.}
Previous results indicate that PEs have limited influence on capturing temporal information for the video modality, prompting a key question: if not positional encodings, what enables VideoLMs to perform effective temporal understanding? We hypothesize that temporal understanding is primarily driven by the order of input video frames, rather than from explicit positional signals. To test this, we compare two interventions applied across all layers during inference: (1) Reverse PE: we reverse the temporal axis of the position IDs while keeping the input frame order unchanged; and (2) Reverse Order: we reverse the input frame order while preserving the original position IDs. For example, given input frames 1 to $N$ with temporal position IDs 1 to $N$, ``Reverse PE" uses frames 1 to $N$ with position IDs $N$ to 1, while ``Reverse Order" uses frames $N$ to 1 with position IDs $N$ to 1.

\subsubsection{Result}
As shown in Figure~\ref{fig:ablate}, reversing the position IDs results in only a modest drop in the predicted ground-truth probability, with changes ranging from -1.13\% to -1.76\% across task types. In contrast, reversing the input frame order leads to a significant performance decline. The largest drop is observed in the ``Multiple Choice" task, with a decrease of 15.45\%, and consistent trends are observed across the remaining task types. \textbf{These results demonstrate that the order of video frames plays a more crucial role than positional encodings in temporal understanding.}



\section{Temporal Information Emerges from Attention-Based Causal Interactions}
In Section~\ref{sec:PE}, we show that positional encodings have limited influence on temporal understanding, particularly in the video modality. In contrast, altering the order of input frames has a more pronounced effect. These findings suggest that temporal information is not primarily captured through explicit positional signals, but instead emerges from the model’s processing of sequential input. 

Given the causal architecture of current VideoLMs, we hypothesize that this order sensitivity arises from the attention mechanism, specifically due to the use of causal masking. Within this structure, temporal representations may be constructed progressively across layers as each token aggregates information from earlier positions.

To investigate this hypothesis, we examine how causal attention mechanisms facilitate the step-by-step construction and propagation of temporal information throughout the model. Drawing inspiration from \citet{zhang2025cross}, we adopt a backward tracing approach, analyzing the model from its output back to earlier interactions. We identify where temporal information originates, how it flows between video and text modalities, and validate its causal nature through targeted analysis.

We adapt the attention knockout method~\cite{geva2023dissecting} to the VideoLM setting in order to analyze this process. Specifically, we modify the attention computation in Equation~\ref{eq:2} by incorporating an additional mask $\bm{M}$ that prevents target tokens in the set $\mathcal{T}$ from attending to source tokens in the set $\mathcal{S}$:
\begin{equation}
\label{eq:4}
\bm{\tilde{A}} = \text{Softmax} \left(
\frac{\text{PE}(\bm{Q}) \cdot \text{PE}(\bm{K})^\mathrm{T}} {\sqrt{d_k}}
+ \bm{M}_C + \bm{M} \right) \bm{V},
\end{equation}
\begin{equation}
\label{eq:5}
m_{i,j} =
\begin{cases}
-\infty, & \text{if}\ (i,j) \in \mathcal{T} \times \mathcal{S}, \\
0, & \text{otherwise}.
\end{cases}
\end{equation}
where $m_{i,j}$ denotes the $(i,j)$-th element of the mask matrix $\bm{M}$, controlling whether token $i$ attends to token $j$ in the attention computation. To localize where temporal information emerges and flows, we apply this knockout iteratively across layers, using a sliding window of \(k = 5\) layers to reduce noise and improve stability. 


\begin{figure}
    \centering
    \includegraphics[width=1\columnwidth]{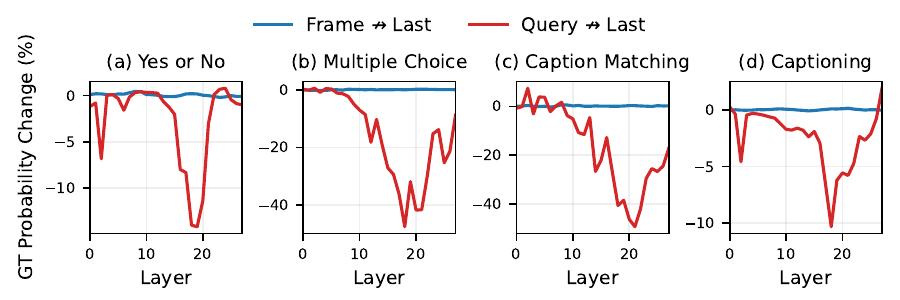}
    \caption{Effect of blocking attention from the final token to the input sources. Disabling query access causes major performance drops, while blocking video tokens has little effect. This highlights the dominant role of query tokens in final-stage temporal understanding.}
    \label{fig:last}
\end{figure}

\subsection{Query-Driven Temporal Prediction}

\subsubsection{Experiment.}
To better understand the full process, we begin by identifying which input tokens directly contribute to the model’s temporal understanding at the output. We perform an attention knockout to the final answer token, blocking its ability to attend to either the query tokens or the video frame tokens. Formally, we set $\mathcal{T} = \{i_{\text{last}}\}$, where $i_{\text{last}}$ is the index of the final output token, and set $\mathcal{S}$ to the indices of either the query tokens or the video tokens, depending on the modality being ablated. This allows us to isolate which modality serves as the immediate source of temporal information during model prediction.

\subsubsection{Result.}

Figure~\ref{fig:last} reveals a clear asymmetry in the effects of blocking attention from the final output token to different input modalities. Disabling access to frame tokens has minimal impact, with ground-truth probabilities remaining near zero across all layers, suggesting that visual features are not directly referenced during final decoding. In contrast, blocking access to query tokens causes substantial and consistent performance drops, particularly in deeper layers, exceeding 40\% in ``Multiple Choice” and ``Caption Matching” tasks. \textbf{These results indicate that temporal understanding at the output stage is predominantly carried by query tokens, which likely synthesize temporal cues accumulated earlier in the model, as further validated in the following sections. In contrast, video tokens play only a minimal direct role in the final prediction.}


\subsection{Stage-wise Temporal Integration}
\begin{figure}
    \centering
    \includegraphics[width=1\columnwidth]{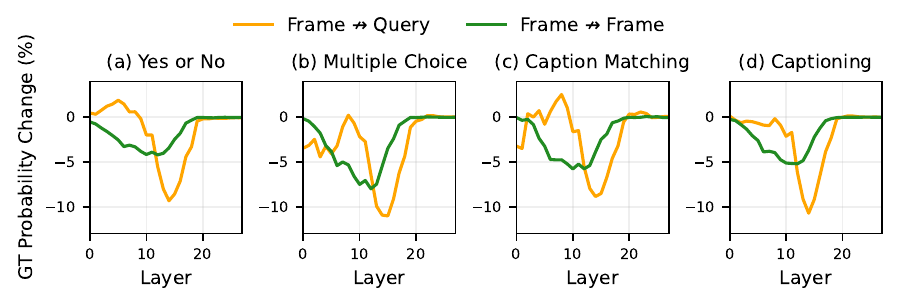}
    \caption{Effect of blocking inter-frame and frame-to-query attention. Probability drops from inter-frame blocking in early layers indicate where temporal information begins to form within the video stream. In contrast, drops resulting from frame-to-query blocking in later layers reveal when this temporal information flows to the query.}
    \label{fig:frame_query}
\end{figure}
\subsubsection{Experiments.}
Although final predictions primarily rely on the query, temporal understanding necessarily requires visual information to recognize patterns of change, motion, and progression over time. To pinpoint when and how video frames contribute to this process, we design two complementary attention knockout setups: (1) Frame-to-query: we prevent query tokens from attending to video frame tokens, (2) Inter-frame: we prevent each video frame from attending to earlier frames. These interventions allow us to isolate two distinct stages in the temporal information pathway: the early construction of temporal relationships through inter-frame attention, and the later integration of this information into the query representation.

\subsubsection{Result.}
As shown in Figure~\ref{fig:frame_query}, the two interventions yield distinct layer-wise effects. Blocking frame-to-query attention leads to a sharp decline in ground-truth probability around layer 15, with the most pronounced drop over 10\% in the ``Captioning'' task. This indicates that the middle layers play a key role in integrating temporal information into the query representation. In contrast, blocking inter-frame attention causes earlier degradation, peaking near layer 10, with probability drops ranging from 4\% to 6\%, suggesting that early layers are critical for forming temporal relationships across frames. \textbf{Together, these observations indicate a two-stage temporal processing process: early layers construct inter-frame temporal relationships, which are then integrated into the query by middle layers to support temporal reasoning in later decoding.}


\subsection{Last-Frame-Centered Temporal Propagation}
\begin{figure}
    \centering
    \includegraphics[width=1\columnwidth]{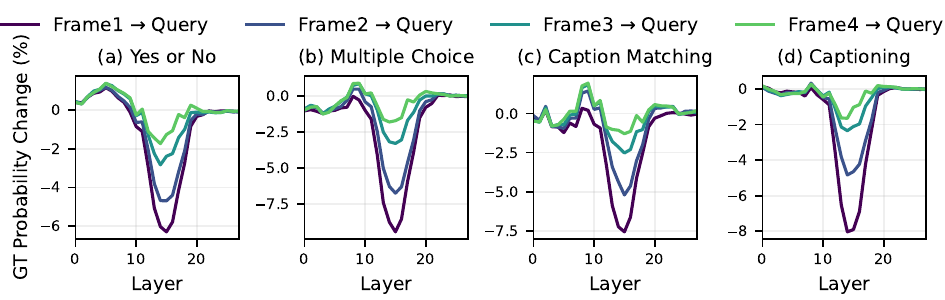}
    \caption{Effect of restricting query attention to single frames. Performance drops sharply when the query can only attend to early frames, but shows minimal drop when attending to later frames. This supports that temporal information flows forward and accumulates in the last frame.}
    \label{fig:frame_num}
\end{figure}

\subsubsection{Experiment.}
Building on the two-stage process where temporal cues are first constructed via inter-frame interaction and then integrated into the query, we now investigate whether certain frames play a more prominent role in transmitting temporal information to the query. To do this, we constrain the model such that the query tokens are only allowed to attend to the visual tokens from a single frame. In our setting, each video contains 4 frames. Formally for each frame $t \in \{1, 2, 3, 4\}$, we enforce this by defining the set $\mathcal{T}$ as all query token indices, and the set $\mathcal{S} = \{i_v \mid i_v \in \mathcal{I}_v \setminus \mathcal{I}_{v_t}\}$, where $\mathcal{I}_v$ is the set of all visual token indices and $\mathcal{I}{v_t}$ denotes those corresponding to frame $t$.

\subsubsection{Result.}
Figure~\ref{fig:frame_num} illustrates the effect of restricting the query’s attention to a single frame. Across all tasks, the largest drop in ground-truth probability occurs when the query attends only to early frames, particularly Frame 1. For instance, in the ``Multiple Choice'' task, this results in a decline of over 8\%, indicating that early frames lack sufficient temporal cues for accurate prediction. In contrast, attending only to later frames, especially Frame 4, yields minimal performance loss, typically below 2\%. This suggests that temporal information accumulates across frames, with the final frame serving as an aggregation point. As a result, the query performs well when attending solely to the last frame, but suffers when limited to early ones. \textbf{Overall, rather than evenly extracting and comparing information for temporal understanding, the query primarily accesses aggregated temporal cues by attending to the final, information-rich frame.}



\subsection{Beyond Correlation to Causality}

\begin{figure}
    \centering
    \includegraphics[width=1\columnwidth]{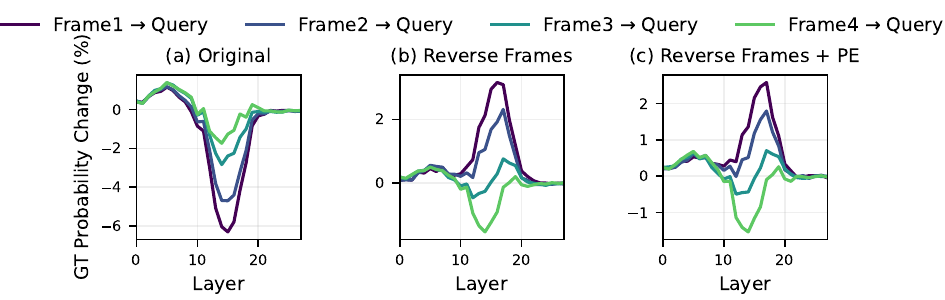}
    \caption{Effect of reversing video order in the “Yes or No” task. Attribution patterns consistently flip regardless of position IDs, suggesting that the model captures temporal information primarily through causal inference, instead of simple visual content collection.}
    \label{fig:reverse}
\end{figure}

\subsubsection{Experiment.}
The previous experiment suggests that the last frame serves as an aggregation point for temporal cues. However, it remains unclear whether this aggregation results from causal processing or merely from content accumulation regardless of frame order. To probe this, we leverage the “Yes or No” task, where reversing the video order flips the correct answer. We repeat the single-frame attribution analysis on reversed sequences to examine whether the model exhibits causal aggregation. If the model simply aggregates visual information, the query should still favor the last frame, as it continues to contain the most information. To further rule out PEs as a confounding factor, we also reverse both the frame order and the position IDs, ensuring that positional signals remain aligned with the original frame sequence.

\subsubsection{Result.}
As shown in Figure~\ref{fig:reverse}, reversing the video frame order leads to a complete inversion of the attribution pattern observed in the previous experiment. Only attending to the initial frame, which previously had the most negative impact, now yields the most positive contribution, even increasing the ground-truth probability. Conversely, the last frame, which had minimal influence before, now causes the greatest drop, despite containing the most visual information. This inversion persists even when position IDs are also reversed, restoring the original positional alignment. \textbf{These observations confirm that the model performs order-sensitive inference through the causal attention mechanism, rather than merely accumulating information.}


\subsection{Spatiotemporal Assembly across Frames}

\subsubsection{Experiments.}
In the last experiment, we move beyond analyzing when and how temporal information flows to the query, and instead investigate how it is causally constructed across frames. Specifically, we examine how later visual content interacts with earlier frames to capture temporal dynamics. Using the prior setup where the query attends only to Frame 4 as a baseline, we isolate how temporal information is encoded within the final frame. We compare three attention configurations to assess the inter-frame construction of temporal information from a spatiotemporal perspective: (1) Corresponding Area: for each frame, attention is restricted to spatially aligned regions in all preceding frames, meaning that each token only attends to tokens located at the same or neighboring spatial positions. As a result, the final frame attends to approximately 300 tokens, while earlier frames attend to fewer; (2) Previous Frame: each frame attends to the immediately preceding frame with about 300 tokens; and (3) Corresponding Area in Previous Frame: each frame attends to the spatially aligned region in the immediately preceding frame with 100 tokens.

\subsubsection{Result.}
\begin{figure}
    \centering
    \includegraphics[width=1\columnwidth]{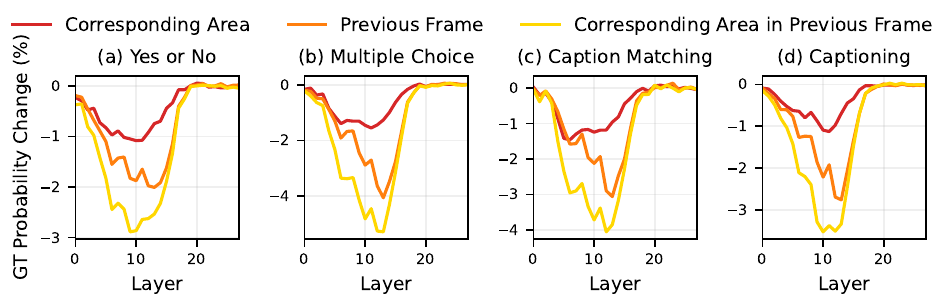}
    \caption{Effect of restricting each frame’s attention to examine inter-frame spatiotemporal contributions to temporal understanding. Sparse long-range attention performs better than dense short-range attention. Early layers prefer global spatial context, and later layers focus on local detail.}
    \label{fig:area}
\end{figure}

Figure~\ref{fig:area} shows that ``Corresponding Area" yields the most stable performance, with ground-truth probability drops consistently under 1\% across all layers. Surprisingly, ``Previous Frame" leads to a larger drop, despite involving more token interactions altogether. Moreover, when comparing ``Previous Frame" and ``Corresponding Area in Previous Frame" with equivalent temporal coverage, early layers exhibit a preference for global spatial context. However, this advantage fades in deeper layers, indicating a transition toward local detail processing. \textbf{Overall, sparse long-range attention process inter-frame temporal integration better than dense short-range interaction, while spatial processing shifts from global to local focus over layers.}


\section{Temporal Pathway-Guided Inference}
Based on our findings that temporal information in VideoLMs arises from a causal pathway, starting from long-range inter-frame interactions and flowing into the query via specific multimodal interaction, we propose two inference strategies that improve efficiency. Both reduce computational and memory costs by pruning token interactions that contribute little to the final temporal representation:

\begin{itemize}
    \item \textbf{Staged Modality Interaction for Sparse Attention}: Temporal information in VideoLMs is progressively constructed, with different layers contributing in distinct ways to the final prediction. Motivated by this observation, we design sparse attention strategies that selectively disable token interactions with limited impact on temporal modeling. For example, (1) in middle layers, query tokens can be restricted to attend only to the last video frame, and (2) in deeper layers, inter-frame interactions can be reduced. This staged sparsification lowers FLOPs while maintaining performance, providing a scalable and effective solution for efficient inference.
    
    \item \textbf{Temporal Exit for KV Cache Compression}: During inference, the key-value (KV) cache grows with sequence length, becoming a major memory bottleneck. However, our analysis shows that not all tokens remain relevant throughout all layers. Leveraging this insight, we can design pruning strategies that remove tokens from the KV cache once their contribution to the final prediction diminishes. For instance, (3) in deeper layers, the large number of frame tokens can be safely discarded. This reduces memory usage without sacrificing performance, providing an effective and compatible solution for improving efficiency in autoregressive generation.
\end{itemize}

To evaluate the effectiveness of our proposed strategies, we conduct validation experiments on the TempCompass and NExT-QA datasets using Qwen-VL-Chat-2.5-7B. The implementation details for the three strategies are as follows: (1) each query token is restricted to attend only to the last frame token in layers 10–20, (2) attention between frame tokens is disabled in layers 20–28, and (3) all frame tokens are removed from the KV cache in layers 20–28.

As shown in Table~\ref{tab:results}, strategies (2) and (3) achieve performance comparable to the baseline across all task types, with accuracy differences typically within 0.2\%. In some cases, they even slightly outperform the baseline. This indicates that both computation and memory can be reduced without sacrificing performance. In contrast, strategy (1), which restricts query attention to only the last frame, leads to a slightly larger accuracy decline. Nonetheless, it remains promising and may benefit from further refinement based on our insightful findings.

\begin{table}[t]
\small
\centering
\begin{tabular}{
    >{\centering\arraybackslash}p{1.1cm} 
    >{\centering\arraybackslash}p{0.85cm} 
    >{\centering\arraybackslash}p{0.85cm} 
    >{\centering\arraybackslash}p{0.85cm} 
    >{\centering\arraybackslash}p{0.85cm}
    >{\centering\arraybackslash}p{1.2cm}
}
    \toprule
    \multirow{2}{*}{\textbf{Method}} & \multicolumn{4}{c}{\textbf{TempCompass}} & \multirow{2}{*}{\textbf{NExT-QA}} \\
    \cmidrule(lr){2-5}
    & Yes/No & MCQ & Caption & Match &  \\
    \midrule
    Baseline & 70.8 & 66.5 & 59.0 & 57.4 & 75.1 \\
    (1)  & 67.9 & 64.1 & 57.6 & 55.0 & 71.9 \\
    (2)  & 70.5 & 66.5 & 59.3 & 57.4 & 75.2 \\
    (3)  & 70.7 & 66.6 & 59.0 & 57.2 & 75.1 \\
    
    \bottomrule
\end{tabular}
\caption{Accuracy on TempCompass and NExT-QA under different efficiency strategies. Strategies (2) and (3) preserve performance while reducing computational or memory cost. Strategy (1) introduces a slight but acceptable drop, with potential for further optimization.}
\label{tab:results}
\end{table}

\section{Conclusion}
This work investigates how modern VideoLMs achieve temporal understanding, a fundamental capability that remains insufficiently explored. While PEs are commonly assumed to be crucial for encoding temporal representations, our analysis shows they contribute only marginally to capturing temporal information in the video input. In contrast, reversing the video sequence while preserving positional signals aligned with the original order results in a significant drop in performance, suggesting that temporal information arises elsewhere. Our findings reveal that temporal understanding emerges from the causal attention mechanism’s order-sensitive structure. Specifically, temporal information is constructed through a causal pathway: it is causally aggregated across frames, integrated into the final frame, and subsequently refined within the query. Building on this insight, we propose two efficiency-focused strategies and validate their effectiveness on benchmark datasets. We hope this work lays the foundation for better temporal modeling in VideoLMs, clarifies key challenges, and encourages further exploration of their internal mechanisms.

\section*{Acknowledgment}
This research is supported by the NTU Start-Up Grant (\#023284-00001), Singapore, and the MOE AcRF Tier 1 Seed Funding Grant (\#025041-00001, RS37/24).

\bibliography{aaai2026}

\appendix
\twocolumn[
  \begin{center}
    {\LARGE \bf Appendix}
    \vspace{2.5em}
  \end{center}
]

The appendix presents extended results from our analysis using Qwen2.5-VL-3B and LLaVA-OneVision-7B to demonstrate the generalizability of our findings. Additionally, we include results on open-ended questions from the ActivityNet-QA dataset to assess the downstream effectiveness of strategies based on our earlier findings.

\section{Additional Analysis Results}
\subsection{Qwen2.5-VL-3B}

The Qwen2.5-VL-3B model and the reported Qwen2.5-VL-7B in the main text belong to the same model family and share fundamental architectural similarities, particularly in their positional encoding designs and multimodal input processing mechanisms. However, they exhibit distinct structural configurations: while the 7B variant employs a wider but shallower architecture with 28 layers, 28 attention heads, and a hidden size of 3,584, the 3B model adopts a narrower yet deeper structure comprising 36 layers, 16 attention heads, and a 2,048-dimensional hidden space. This structural variation allows us to assess how key findings generalize across different depth–width configurations in video language models, with the present study focusing specifically on the 3B variant to evaluate the robustness of the observed phenomena.

\begin{figure}[H]
    \centering
    \includegraphics[width=1\columnwidth]{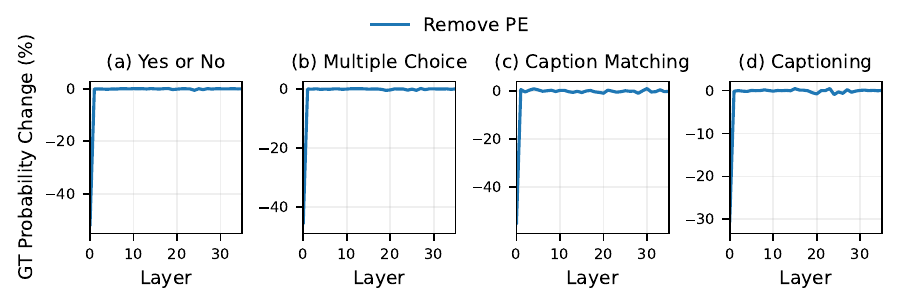}
    \caption{Layer-wise PE ablation effects on Qwen2.5-VL-3B. As with the 7B variant, ablating positional encodings significantly affects only the first layer, with negligible impact on subsequent layers.}
    \label{fig:position_3B}
\end{figure}

\subsubsection{Analyzing the Impact of Positional Encodings.}
We conduct the same positional encoding (PE) ablation analysis on the Qwen2.5-VL-3B model as described in the main text. As shown in Figure~\ref{fig:position_3B}, the impact of removing PEs is largely confined to the first layer, with consistent trends observed across multiple task types. This pattern closely mirrors the behavior seen in the Qwen2.5-VL-7B model. Interestingly, the 3B variant exhibits more uniform performance across task types, possibly due to its deeper architecture facilitating more consistent information propagation and representation learning across layers. These findings suggest that, under different architectural regimes (e.g., deep-narrow vs. shallow-wide), PEs in most layers contribute minimally to temporal understanding.

\begin{figure}[H]
    \centering
    \includegraphics[width=0.85\columnwidth]{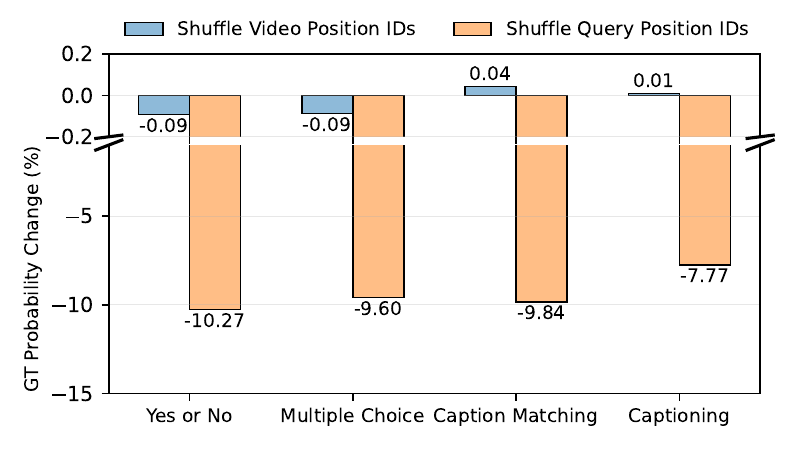}
    \caption{Impact of modality-specific position ID shuffling at the first layer in Qwen2.5-VL-3B. Disrupting positional IDs for textual inputs significantly degrades temporal understanding performance, while doing so for video inputs has minimal effect.}
    \label{fig:shuffle_3B}
\end{figure}

As shown in Figure~\ref{fig:shuffle_3B}, in the modality-specific position ID shuffling experiment, the Qwen2.5-VL-3B model exhibits behavior highly consistent with that of the 7B variant: the observed impact on temporal understanding primarily stems from shuffling PEs of textual inputs, while disrupting video PEs has negligible effects. However, overall, the effect of textual PEs in the 3B model appears weaker than in the 7B model. This difference may be attributed to the deeper architecture of the 3B variant, which could allow greater flexibility in compensating for disrupted positional signals through deeper contextual integration.

\begin{figure}[H]
    \centering
    \includegraphics[width=0.85\columnwidth]{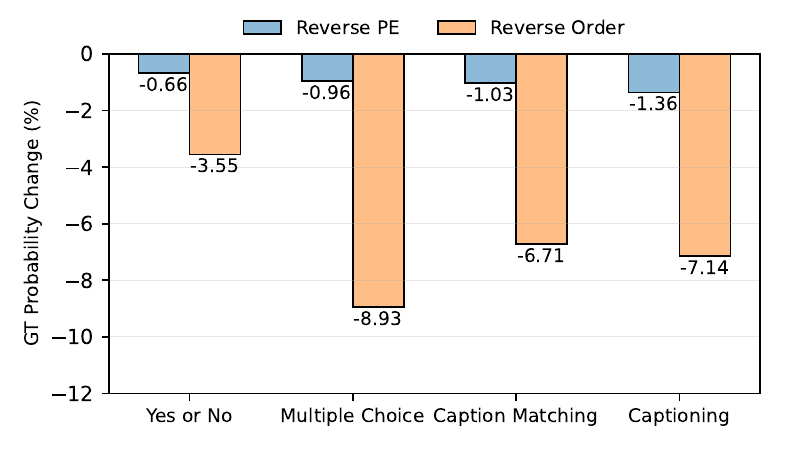}
    \caption{Effect of reversing position IDs versus frame order in Qwen2.5-VL-3B. Reversing the frame order causes a pronounced drop in ground-truth probability, while reversing PEs produces only a minimal impact, further demonstrating their limited effectiveness.}
    \label{fig:ablate_3B}
\end{figure}

In the experiment comparing the effects of ablating PEs and altering frame order, we observe that reversing the position IDs of video frames leads to only a minor drop in ground-truth probability within 1.4\% across all task types in Figure~\ref{fig:ablate_3B}. In contrast, reversing the frame order while preserving the original position IDs results in substantial degradation, with performance drops of up to 8.93\%. These findings underscore the critical role of frame ordering in temporal understanding, while further highlighting the limited contribution of PEs across model structures.

\begin{figure}[H]
    \centering
    \includegraphics[width=1\columnwidth]{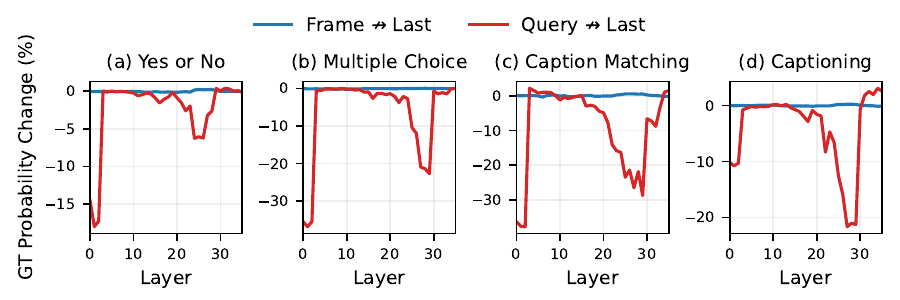}
    \caption{Effect of blocking attention from the final token to different input modalities in Qwen2.5-VL-3B. Restricting attention to query tokens substantially degrades temporal understanding performance, while blocking video tokens has a negligible effect.}
    \label{fig:last_3B}
\end{figure}

\subsubsection{Exploring Causal Temporal Generation Pathway.}
As shown in Figure~\ref{fig:last_3B}, blocking attention from the final prediction token to video tokens in Qwen2.5-VL-3B continues to have minimal impact. In contrast, restricting access to query tokens results in significant performance drops, particularly in deeper layers. Interestingly, unlike the 7B model, the 3B variant also shows sensitivity to query blocking in the first few layers, suggesting that it relies on query tokens earlier in the processing pipeline. This early drop reflects a structural adaptation in the deeper, narrower architecture of the 3B model, which enables earlier layers to more effectively extract and utilize query information. Nonetheless, across both models, query tokens remain the dominant source of temporal cues for final-stage prediction.

\begin{figure}[H]
    \centering
    \includegraphics[width=1\columnwidth]{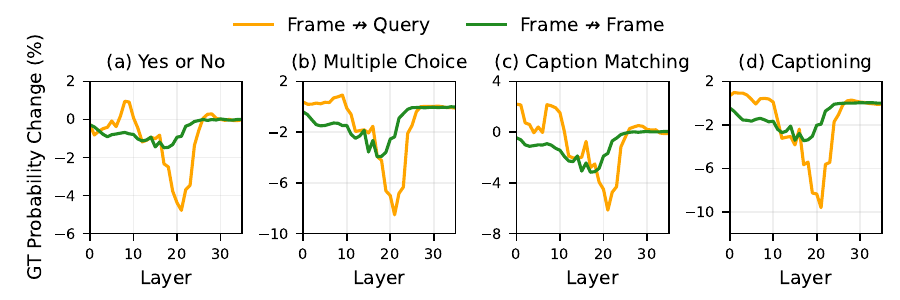}
    \caption{Effect of blocking inter-frame and frame-to-query attention in Qwen2.5-VL-3B. Inter-frame blocking causes early ground-truth probability drops, marking temporal encoding onset; frame-to-query blocking causes later drops, showing information flow to the query.}
    \label{fig:query_3B}
\end{figure}

Figure~\ref{fig:query_3B} shows that Qwen2.5-VL-3B also exhibits a two-stage temporal processing pattern: early drops from inter-frame attention blocking indicate where temporal features are composed across frames, while later drops from frame-to-query blocking reveal when this information flows to the query. Compared to the 7B model, the inter-frame effects in 3B are notably milder, suggesting that the deeper architecture of the 3B model may be less effective at capturing temporal information through inter-frame interactions, an observation consistent with intuition.

\begin{figure}[H]
    \centering
    \includegraphics[width=1\columnwidth]{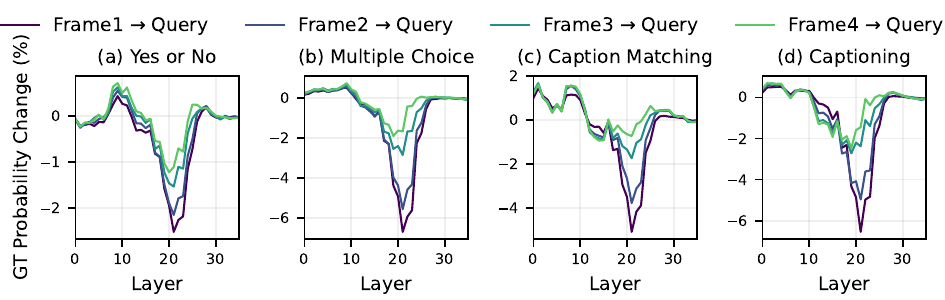}
    \caption{Effect of restricting query attention to single frames in Qwen2.5-VL-3B. Attending only to early frames leads to greater performance drops, suggesting limited temporal information accumulated at those points; minimal impact when attending to the last frame indicates that temporal information has largely converged there.}
    \label{fig:num_3B}
\end{figure}

In Qwen2.5-VL-3B, the final frame also appears to aggregate most of the temporal information, as Figure~\ref{fig:num_3B} shows that restricting query attention to the last frame in middle layers results in minimal performance drop, typically under 2\%. However, the differences between attending to earlier frames (e.g., Frame 1 vs. Frame 2) are less pronounced compared to the 7B model, suggesting that Qwen2.5-VL-3B has a weaker capacity for inter-frame temporal aggregation, likely due to its smaller model size.

\begin{figure}[H]
    \centering
    \includegraphics[width=1\columnwidth]{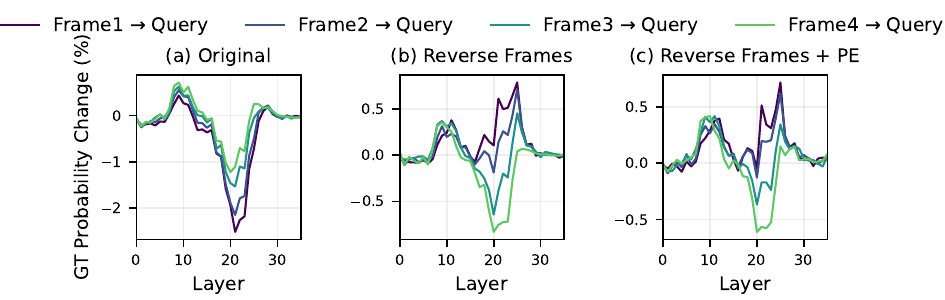}
    \caption{Effect of reversing video order in the ``Yes or No” task in Qwen2.5-VL-3B. Attribution patterns invert, suggesting that temporal understanding depends on causal processing rather than simple visual accumulation.}
    \label{fig:reverse_3B}
\end{figure}

As shown in Figure~\ref{fig:reverse_3B}, for the ``Yes or No" task, reversing the frame order completely flips the attribution: forcing the query to attend to the last frame now yields the worst performance, even when the position IDs still follow the original frame order. This indicates that temporal information in the 3B model is also aggregated in a causal, order-sensitive manner, rather than through simple accumulation.

\begin{figure}[H]
    \centering
    \includegraphics[width=1\columnwidth]{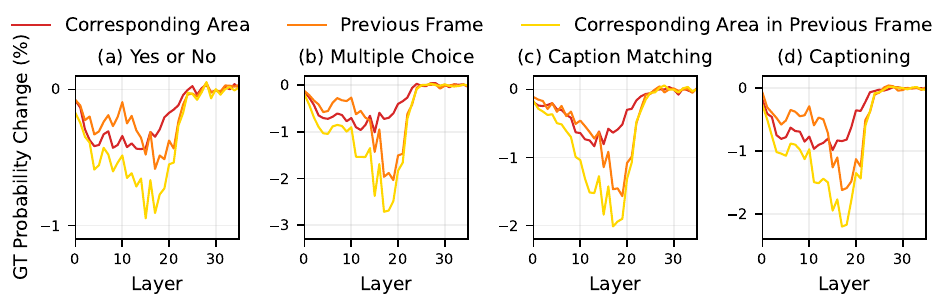}
    \caption{Effect of restricting attention to analyze inter-frame spatiotemporal contributions in Qwen2.5-VL-3B. Sparse, long-range attention outperforms dense, local attention. Spatially, early layers favor global context, while later layers attend to finer details.}
    \label{fig:area_3B}
\end{figure}

Qwen2.5-VL-3B and the 7B model share a similar process for constructing inter-frame temporal information. As shown in Figure~\ref{fig:area_3B}, restricting each frame to attend only to corresponding regions across all previous frames yields better performance than attending to the entire previous frame, despite involving fewer total token interactions. From a temporal perspective, the length of context appears more important than its density. Spatially, the difference between attending to the full previous frame and its corresponding region is evident only in the early and middle layers, with little distinction in deeper layers. This suggests that the model initially relies on global spatial context and gradually shifts its focus to finer local details as depth increases.

\subsection{LLaVA-OneVision-7B}

LLaVA-OneVision-7B differs from the Qwen2.5-VL model series in several architectural aspects. For instance, it does not merge video frame features. However, a more critical distinction is its lack of any dedicated positional encoding design for video inputs. Instead, it directly inherits the rotary positional encoding from its base language model, applying simple 1D position IDs sequentially across video tokens, without explicitly modeling frame boundaries or spatial structure. The use of this standard 1D encoding, which has proven effective in large language models, helps eliminate the possibility that the ineffectiveness of positional encodings is due to overly complex or poorly tuned design choices. Therefore, this minimal design makes LLaVA-OneVision-7B a strong candidate for evaluating the generalizability and validity of our findings on positional encodings.

\begin{figure}[H]
    \centering
    \includegraphics[width=1\columnwidth]{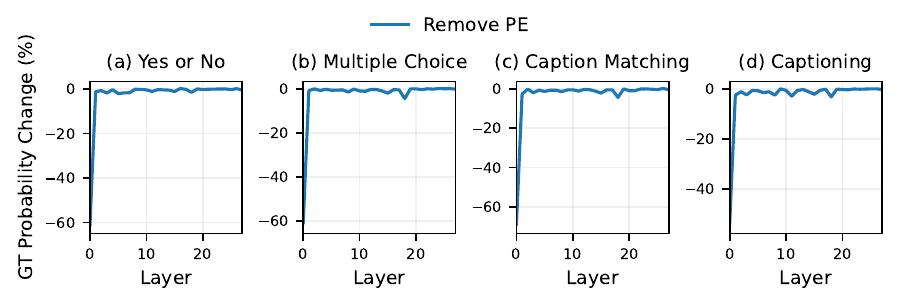}
    \caption{Layer-specific effects of removing positional encodings in LLaVA-OneVision-7B. The removal of positional encodings only has a strong effect on the first layer but minimal impact on later layers.}
    \label{fig:position_llava}
\end{figure}

As shown in Figure~\ref{fig:position_llava}, the layer-wise PE ablation results on LLaVA-OneVision-7B follow the same trend observed in Qwen2.5-VL models. Removing PEs in the first layer leads to a substantial drop in the prediction probability of the ground-truth, exceeding 60\% in nearly all task types. In contrast, removing PEs from subsequent layers results in negligible performance change, with the ground-truth probability curve remaining nearly flat throughout. This consistent pattern, despite architectural and design differences across models, suggests that positional encodings primarily contribute to temporal understanding only in the first layer, while later layers rely more on learned contextual features rather than explicit position signals.

\begin{figure}[H]
    \centering
    \includegraphics[width=0.85\columnwidth]{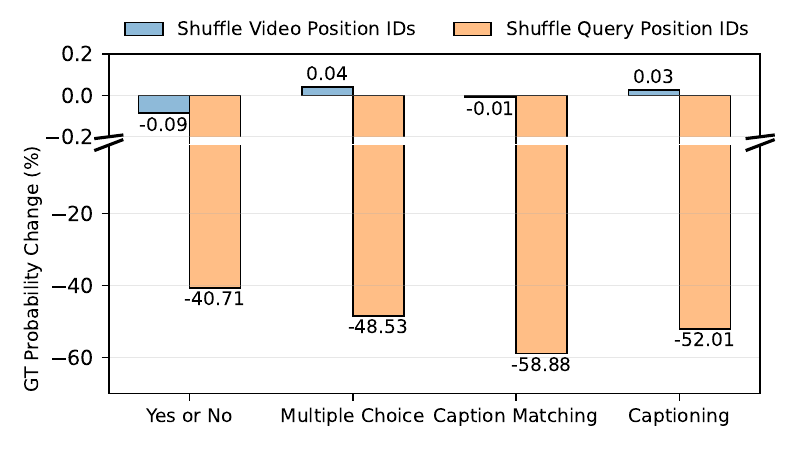}
    \caption{Effect of modality-specific position ID shuffling at the first layer in LLaVA-OneVision-7B. Altering positional IDs for textual inputs leads to a significant drop in temporal understanding performance, whereas a similar disruption for video inputs has little impact.}
    \label{fig:shuffle_llava}
\end{figure}

As shown in Figure~\ref{fig:shuffle_llava}, LLaVA-OneVision-7B exhibits a similar trend in the modality-specific position ID shuffling experiment. Shuffling the position IDs of textual tokens at the first layer results in a substantial drop in ground-truth prediction probability across all tasks, up to 58.88\% for ``Caption Matching" task. In contrast, shuffling the position IDs of video tokens has minimal impact, with performance changes remaining below 0.1\%. These results indicate that even simple and effective 1D positional encodings, when applied to video inputs, still fail to contribute meaningfully to temporal information capture.

\begin{figure}[H]
    \centering
    \includegraphics[width=0.85\columnwidth]{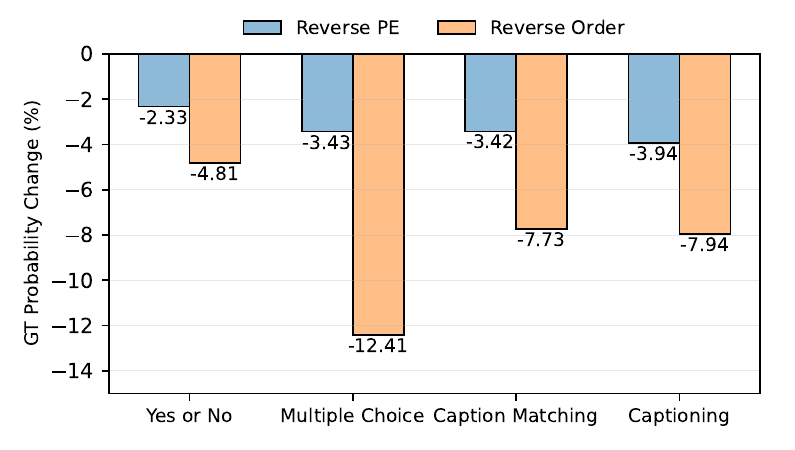}
    \caption{Effect of reversing position IDs versus frame order in LLaVA-OneVision-7B. Reversing frame order still causes a clear drop, while reversing position IDs has minimal effect. Compared to Qwen2.5-VL, the gap is smaller, suggesting that simpler designs may be more easily utilized.}
    \label{fig:ablate_llava}
\end{figure}

We further evaluate the role of PEs in temporal modeling by comparing the effects of reversing frame order versus reversing position IDs in LLaVA-OneVision-7B. As shown in Figure~\ref{fig:ablate_llava}, reversing the frame order leads to a noticeable drop in ground-truth prediction probability, indicating that the model is still sensitive to the frame order. In contrast, reversing the position IDs results in only minor performance degradation, with changes mostly within 4\%. This trend aligns with findings in Qwen2.5-VL models, reaffirming the limited impact of PEs. However, the performance gap between the two conditions is smaller in LLaVA-OneVision-7B, suggesting that its simpler PE design may be more easily exploited by the model.

\section{Additional Downstream Task Verification}
We evaluate three proposed efficiency strategies on the ActivityNet-QA dataset using the Qwen2.5-VL-7B model. The strategies are as follows: (1) between layers 10–20, restrict the query’s attention to only the final frame in order to reduce computation; (2) disable inter-frame attention across layers 20–28 to reduce computation; and (3) completely discard all frame tokens after layer 20 to reduce memory usage. ActivityNet-QA is an open-ended question-answering task, so we allow the model to generate responses of up to 100 tokens per question. The quality of generated answers is assessed using GPT-4o-mini as the evaluator.

\begin{table}[H]
\small
\centering
\begin{tabular}{
    >{\centering\arraybackslash}p{1.6cm} 
    >{\centering\arraybackslash}p{1cm} 
    >{\centering\arraybackslash}p{1cm} 
    >{\centering\arraybackslash}p{1cm} 
    >{\centering\arraybackslash}p{1cm}
}
    \toprule
    \textbf{Metric} & \textbf{Baseline} & \textbf{(1)} & \textbf{(2)} & \textbf{(3)} \\
    \midrule
    Accuracy & 46.9 & 40.4 & 46.9 & 42.7 \\
    Score    & 2.71 & 2.39 & 2.71 & 2.54 \\
    \bottomrule
\end{tabular}
\caption{Accuracy and score on ActivityNet-QA across different efficient inference strategies based on our findings.}
\label{tab:results}
\end{table}

As shown in Table~\ref{tab:results}, Strategy (2), which disables inter-frame attention between layers 20–28, achieves the same accuracy (46.9\%) and score (2.71) as the full baseline model, demonstrating that later-stage inter-frame interactions can be safely pruned without sacrificing any performance. Strategy (3), which discards all frame tokens after layer 20, shows a moderate drop in accuracy (42.7\%), suggesting that much of the temporal understanding is already completed in earlier layers. It is worth noting that ActivityNet-QA is a more challenging dataset, with longer video durations and a stronger reliance on temporal information. Strategy (1), which limits query attention to only the final frame between layers 10–20, results in a more noticeable performance degradation. Despite being relatively aggressive, the strategy maintains over 40\% accuracy, suggesting potential for further optimization to reduce performance loss. Overall, these results validate the practicality and insightfulness of our findings and highlight their potential for enabling more efficient model designs.

\end{document}